%% file: main.tex
\pdfoutput=1

\documentclass[11pt]{article}

\usepackage{EMNLP2023}
\usepackage{balance}
\usepackage{times}
\usepackage{latexsym}

\usepackage[T1]{fontenc}

\usepackage[utf8]{inputenc}

\usepackage{microtype}

\usepackage{inconsolata}

\usepackage{graphicx}
\usepackage{subfigure}
\usepackage{booktabs, multirow} 
\usepackage{hyperref}

\usepackage{amsmath}
\usepackage{amssymb}
\usepackage{mathtools}
\usepackage{amsthm}
\usepackage{hyperref}
\usepackage{balance}
\usepackage{tablefootnote} 

\newcommand{\qidn}{\textrm{Q}_\textrm{id}}

\newcommand{\sqid}{\mathcal{T}_{\mathtt{q}}}
\newcommand{\tqid}{\mathcal{G}_{\mathtt{q}}}

\usepackage[capitalize,noabbrev]{cleveref}

\usepackage[utf8]{inputenc}
\theoremstyle{plain}

\theoremstyle{definition}

\theoremstyle{remark}

\usepackage[textsize=tiny]{todonotes}

\title{Construction of Paired Knowledge Graph - Text Datasets \\Informed by Cyclic Evaluation}

\author{Ali Mousavi$^{*\dagger}$, Xin Zhan$^{*\dagger}$, He Bai$^{*\dagger}$, Peng Shi$^{\ddagger}$, Theo Rekatsinas$^{\dagger}$, Benjamin Han$^{\dagger}$,\\
        {\bf Yunyao Li$^{\dagger}$, Jeff Pound$^{\dagger}$, Josh Susskind$^{\dagger}$, Natalie Schluter$^{\dagger}$, Ihab Ilyas$^{\dagger}$, Navdeep Jaitly$^{\dagger}$} \\
        $^{\dagger}$Apple Inc,  $^{\ddagger}$ University of Waterloo}

\begin{document}
\maketitle
\def\thefootnote{*}\footnotetext{These authors contributed equally to this work}\def\thefootnote{\arabic{footnote}}

\input{Sections/abstract_v2}

\input{Sections/intro_v2.tex}
\input{Sections/related_work_v2.tex}

\input{Sections/cyclic_eval.tex}
\input{Sections/datasets.tex}

\input{Sections/experiments.tex}

\input{Sections/conclusion}
\input{Sections/limitations.tex}
\input{Sections/ethics.tex}


\bibliography{anthology,references}
\bibliographystyle{acl_natbib}

\input{Sections/appendix.tex}


\end{document}

%% file: Sections/abstract_v2.tex
\begin{abstract}
Datasets that pair Knowledge Graphs (KG) and text together (KG-T) can be used to train forward and reverse neural models that generate text from KG and vice versa.
However models trained on datasets where KG and text pairs are not equivalent can suffer from more hallucination and poorer recall.
In this paper, we verify this empirically by generating datasets with different levels of noise and find that noisier datasets do indeed lead to more hallucination.
We argue that the ability of forward and reverse models trained on a dataset to cyclically regenerate source KG or text is a proxy for the equivalence between the KG and the text in the dataset.
Using cyclic evaluation we find that manually created WebNLG is much better than automatically created TeKGen and T-REx.
Guided by these observations, we construct a new, improved dataset called \textbf{LAGRANGE} using heuristics meant to improve equivalence between KG and text and show the impact of each of the heuristics on cyclic evaluation.
We also construct two synthetic datasets using large language models (LLMs), and observe that these are conducive to models that perform significantly well on cyclic generation of text, but less so on cyclic generation of KGs\footnote{We will release all the datasets constructed.}, probably because of a lack of a consistent underlying ontology.

\end{abstract}

%% file: Sections/intro_v2.tex
\section{Introduction}\label{sec:intro}
The Natural Language Processing community has recently released several datasets with paired knowledge graphs (KG) and associated text (which we will refer to as KG-T) such as WebNLG~\cite{webnlg}, TeKGen~\citep{tekgen}, KGPT~\citep{kgpt} and T-REx~\cite{trex}.
Such datasets can be used to train sequence-to-sequence models that can generate text from KGs (forward model) or vice versa (reverse model).  
However, prior studies assert that sequence-to-sequence models learn to hallucinate when the conditioning data has poor correlation with the sequence being produced, which can be the case when training data is noisy~\cite{hallucination}.
In KG-text domain, hallucination can be quite problematic when the goal is to generate factually correct statements from KGs, in scenarios such as Question Answering.

\begin{table}[!t]\centering
\small
\begin{tabular}{lcccc}\toprule
\multirow{2}{*}{Dataset} &\multicolumn{2}{c}{Graph-to-Text} &\multicolumn{2}{c}{Text-to-Graph} \\\cmidrule{2-5}
&BLEU-4 &{\hskip -0.1in}ROUGE-4 &Precision &{\hskip -0.1in}Recall \\\midrule
WebNLG &44.59 &31.30 &90.00 &89.40 \\
+10\% noise &44.46 &31.44 &89.79 &88.76 \\
+20\% noise &43.97 &30.96 &89.43 &88.29 \\
+30\% noise &43.54 &30.54 &88.16 &87.28 \\
+40\% noise &42.56 &29.92 &87.05 &86.83 \\
+50\% noise &41.56 &28.73 &83.65 &85.51 \\
\bottomrule
\end{tabular}
\caption{Evaluation of models trained with noisy data.}\label{tab:webnlg-noise-unidirection-eval}
\end{table}

When a KG-T evaluation dataset is available, it is easy to assess hallucination and recall of models trained on the data. 
For forward models,  BLEU score between the text generated from the KG and the ground truth can be seen as a proxy for hallucination, while ROUGE score can be seen as a proxy for recall.
For reverse models, comparing KG generated from the text, with the ground truth reveals how many KG facts are hallucinated, and how many are recalled. 
In Table~\ref{tab:webnlg-noise-unidirection-eval} we show that as more and more noise is added to the KG part of WebNLG, which is manually created, the quality of text generated by forward models trained on it deteriorates, and so does the quality of the KG generated by the reverse models.
Thus, a KG-T dataset which can be used to train reliable forward and reverse models needs to have as less noise as possible in the triples, and further, the information content between the text and the KG needs to be similar, as is the case with WebNLG.

However, most automatically generated datasets such as KGPT, TeKGen, T-REx have relatively sparse coverage of text with KG, since they are derived from existing KG datasets like Wikidata, whose coverage is relatively sparse.
In these cases not only do models trained on these datasets hallucinate more, it is also hard to assess their accuracy on held out validation sets\footnote{Which are also automatically created.} because the validation sets themselves are highly noisy. Therefore, deciding on the best dataset for training these models can be challenging due to various factors, such as the peculiarities of the data KG ontology, the types of sentences found in different datasets, and so on. This makes it difficult to compare the results effectively.


In this paper, we claim that cyclic generation is a meaningful way of assessing the hallucination and recall of neural models trained on KG-T datasets, when a manually labelled set that has a comprehensive coverage of the text with KG is unavailable.
In cyclic generation we start from one side (text or KG) and generate its counterpart (KG or text respectively) using the appropriate (forward or reverse) sequence-to-sequence model trained on a KG-T dataset.
The source is then regenerated using the model that works in the opposite direction.
When we start from the graph, we call the cyclic reconstruction GTG; and when we start from the text, we call it TGT.
GTG measures the ability to reproduce the KG with its specific ontological requirements while TGT measures the ability to reproduce data in more free formed text.
Again, BLEU score between the original text and the reconstruction in TGT can be seen as a measure of hallucination, while the ROUGE score can be seen as a measure of recall. In GTG, triples can be matched more reliably than text to measure precision and recall of the facts, since the triples follow an ontology and comparison is less ambiguous than comparing free form text.

In these settings, cyclic evaluation is a better way of assessing which dataset to train the neural models, compared to assessing forward and reverse models separately, unidirectionally. This is because the evaluation does not rely on knowing ground truth matches, which are unavailable because the datasets are automatically constructed by alignment.
Instead, it can rely on the sentences from the datasets, or the KG alone, separately.
This is reminiscent of back-translation as being a way of assessing the quality of machine translation -- since the assessment is performed on the known ground truth itself.

We use this method to compare several KG-T datasets and show that manually created WebNLG is much better than TeKGen and T-REx which are constructed automatically by aligning Wikipedia sentences with Wikidata.
We use the lessons learnt to construct a new, \textbf{la}rge-scale \textbf{gra}ph-text alig\textbf{n}ed dataset for graph-text cross-modal \textbf{ge}neration (\textbf{LAGRANGE}) using heuristics meant to improve alignment and coverage and show how each of the heuristics improves cyclic generation.
We also construct two synthetic datasets using large language models (LLMs), and observe that these produce models that do very well on cyclic generation of text, but less so on cyclic generation of KGs. 
We hypothesize that this is probably because they lack a consistent ontology from one example to the next, which makes it difficult for neural models to reconstruct the exact KG through cyclic generation. 
This is meanwhile, not a problem for generating the text cyclically, since the neural network models are able to learn to deal with the variability in ontology, when reconstructing text from the ``KG'' of the dataset generated by the LLMs.


%% file: Sections/related_work_v2.tex
\section{Related Work}\label{sec:related_work}
Prior surveys have reported on the impact of noisy data on hallucination~\cite{hallucination} in sequence to sequence models, calling it ``source-reference divergence''. 
Other works have tried to reduce hallucination, for example by penalizing outputs that are hallucinations~\cite{zhou2021detecting}. 
The use of cyclic generation is not new -- it has been used as a way of improving generative models in KG-text settings \citep{wang2023faithful,guo2020cyclegt}. 
In contrast, we claim that a part of the reason why cyclic generation is poor, is because poor equivalence between KG and text in the dataset teaches the model to hallucinate missing facts.  
We thus propose to evaluate the quality of aligned graph-text datasets by measuring the cyclic generation abilities of models trained on them.

We briefly describe how our approach to create LAGRANGE is different from how other KG-T datasets were created. 
WebNLG \citep{webnlg}) is a small scale manually created dataset and is thus of high quality but has a limited ontology. 
KGPT\cite{kgpt}, GenWiki \citep{jin2020genwiki} and TeKGen\citep{tekgen} align Wikidata triples to the text in Wikipedia articles, by having different strategies for matching subjects or objects and hyperlinks in the text to Wikidata triples, but these methods do not check for semantic relevance of the KG to the sentence. 
In contrast, T-REx~\cite{trex} utilizes predicate linker and coreference resolution to match KG triples to text, but may miss matches when the predicate is semantically entailed but not explicitly mentioned. 
See the Appendix~\ref{sec:appendix} for a more detailed explanation.

%% file: Sections/cyclic_eval.tex
\section{Cyclic Evaluation of KG-T Datasets}\label{sec:align}

\begin{figure}[b]
\hspace{-0.1in}
\centering
\includegraphics[width=0.365\textwidth]{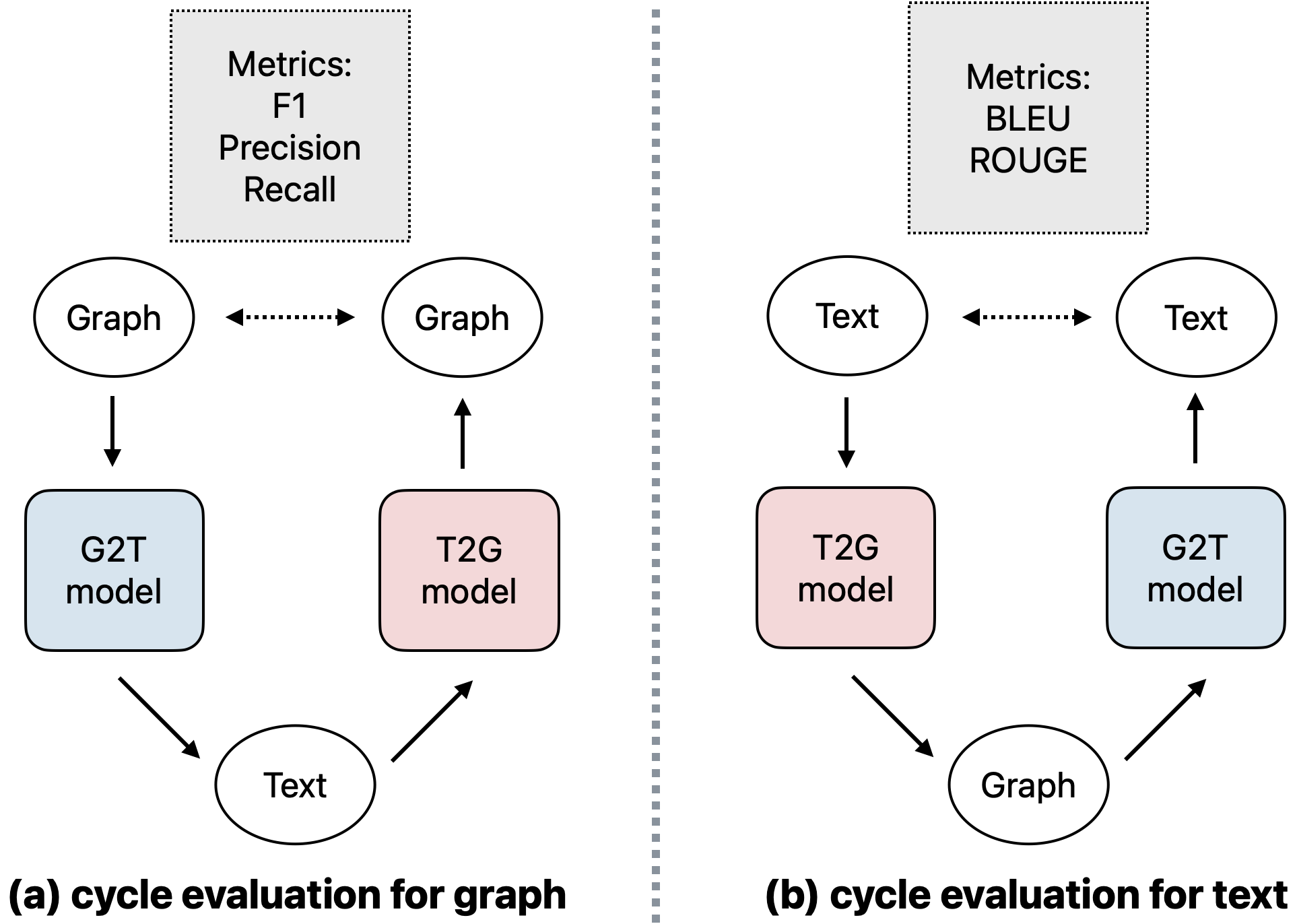}
\caption{The cycle evaluation for dataset alignment.}
\label{fig:cycle}
\end{figure}

A KG-T dataset is defined as a set of $N$ paired (graph, text) tuples, $\{(\mathcal{G}_i, \mathcal{T}_i)\}_{i= 1\cdots N}$ where each graph $\mathcal{G}_i$ is matched to a natural language sentence (or paragraph) $\mathcal{T}_i$. 
Here each graph, $\mathcal{G}_i$, is a set of $K_i$ tuples $\{(s_j, p_j, o_j)\}_{j=1\cdots K_i}$ where each tuple describes a relationship (predicate) $p_j$ between a subject $s_j$ and an object $o_j$. 
\footnote{Note that we sometimes refer to the whole collection of triples in a dataset such as Wikidata as the KG, and the subset that is matched to a particular sentence in a KG-T dataset also as the KG. We use these interchangeably, but it should be obvious from context.} 

We train the parameters $\theta$ of a model $G2T(\cdot; \theta)$ to predict the text $\mathcal{T}$ associated with the graph, $\mathcal{G}$, by minimizing a loss function $l(\cdot,\cdot)$,
\begin{equation}
\min_{\theta} \sum_{i=1}^N l(G2T(\mathcal{G}_i; \theta), \mathcal{T}_i)
\end{equation}
Similarly, we also train a model $T2G(\cdot; \phi)$ to predict the graph $\mathcal{G}$ associated with the text $\mathcal{T}$ by minimizing an appropriate loss function.
The models resemble the characteristics of a KG-T dataset and reflect the degree of hallucination and recall during generation.

For cyclic evaluations (see Figure~\ref{fig:cycle}), we compute a $GTG$ score,  $s(\mathcal{G'}, \mathcal{G})$,  which compares a cyclically generated set of triples $\mathcal{G'}=T2G(G2T(\mathcal{G};\theta);\phi)$ against the original set of triples, $\mathcal{G}$. 
Similarly we compute a $TGT$ evaluation by computing a score $s(\mathcal{T}', \mathcal{T})$ which compares a cyclically generated sentence $\mathcal{T}'=G2T(T2G(\mathcal{T};\phi);\theta)$ against the original $\mathcal{T}$. 


We can use different models, loss functions and scores for the assessment. 
In this paper, we trained transformer based T5 models, with a sequence to sequence loss function. 
The quality of the results was assessed using various metric scores, including BLEU, ROUGE, and others (see section~\ref{sec:exps} for more details).

%% file: Sections/datasets.tex
\section{Datasets}\label{sec:rawdata}
In this section, we describe the methodology used to create LAGRANGE and the synthetic datasets.
\input{Sections/lagrange_v2}
\input{Sections/synthetic.tex}

%% file: Sections/lagrange_v2.tex
\subsection{LAGRANGE}
LAGRANGE consists of pairs of aligned KG triples from Wikidata and sentences from Wikipedia. 
We created an initial alignment between Wikidata KG triples, and Wikipedia using string matching techniques, and subsequently filtered out low quality matches using a semantic entailment model.
Finally we augmented the KG triples by generating from a T2G model.

\subsubsection{Generating an Initial Alignment}\label{subsec:align}

\begin{figure}[b]
\centering
\includegraphics[width=0.45\textwidth]{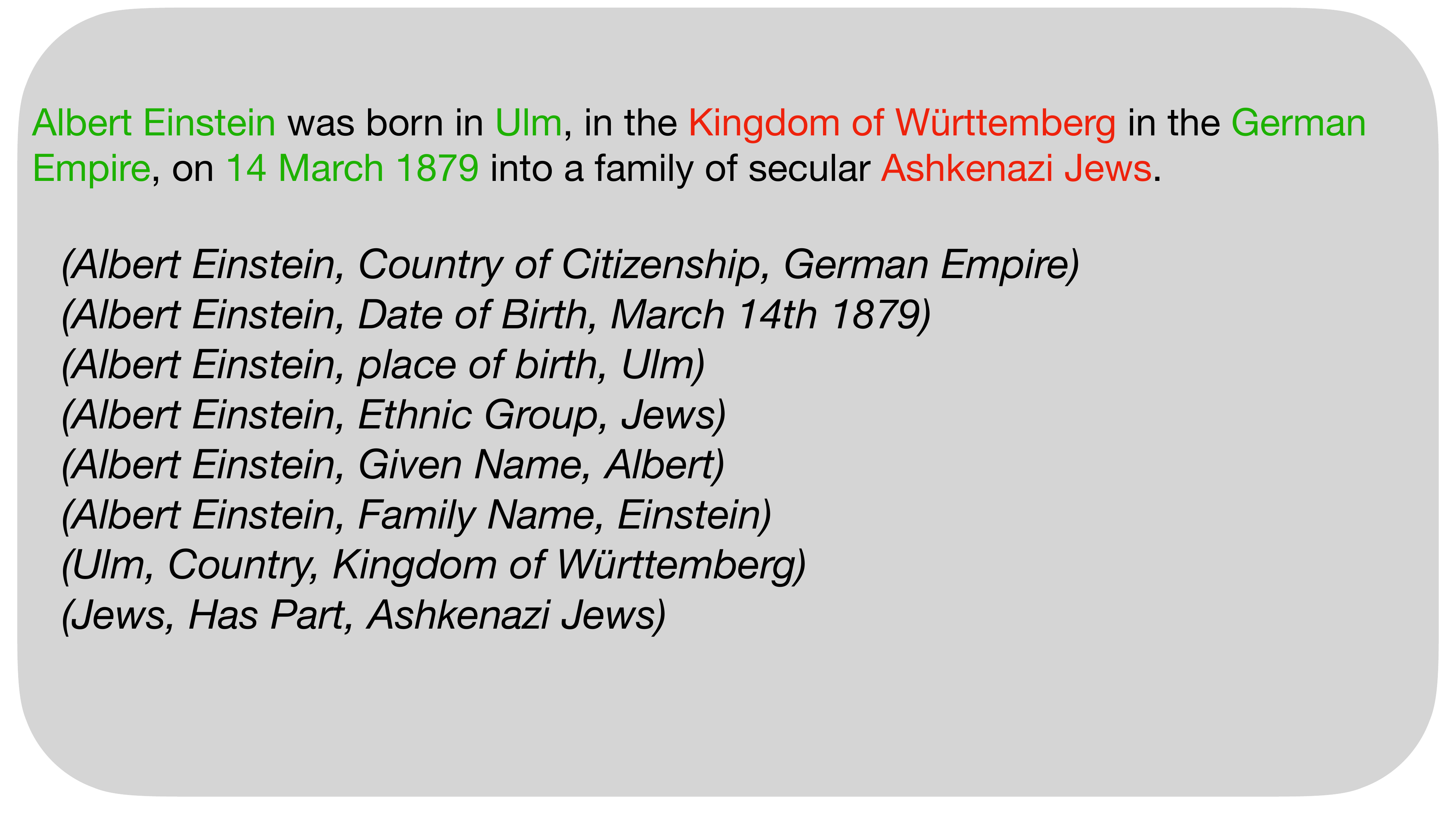}
\caption{An example sentence from the Albert Einstein's Wikipedia article and matched triples to it. Note that our approach goes beyond first-hop neighbors by considering the second-hop neighbors.}
\label{fig:wiki_triples}
\end{figure}

At a broad level, Wikidata can be described as a collection of KG triples $(s, p, o)$, representing a relationship (referred to as a predicate), $p$, between a subject entity, $s$, and an object entity $o$.
An initial pairing between Wikipedia sentences and Wikidata KG triples is easily achieved by matching the subject of the Wikipedia page containing the sentence to Wikidata triples about the same subject\footnote{These are nicely linked together through a unique identifier that Wikidata calls $\qidn$.}. We additionally make sure that the subject or its aliases is explicitly referenced in the sentence.
These initial matches are then filtered to remove KG triples where the object entity, or its alias is not matched to the sentence. 
The remaining triples are regarded as first-hop matches.
Note that the actual dataset construction deals with corner cases of compound predicates in Wikidata, handling of dates and aliases, among other factors. A more detailed and formal description of the construction can be found in the appendix in section~\ref{sec:appendix_initial_alignment} as some of the details are not essential to understanding the main theme of the paper.

\subsubsection{Incorporating Second-Hop Neighbors}\label{subsec:sh}
A significant number of sentences contain additional information in them that does not relate to the subject entity, but to other entities in the sentence.
In order to ensure a good coverage of the information present in the sentence, we also matched second-hop KG triples -- which are triples whose subject is an entity that was an object in one of the triples in the first-hop alignment. 
As before, we ensure that the object entities of these second-hop triples are found in the sentence. See Figure~\ref{fig:wiki_triples} for an example of aligned text and its corresponding KG.

\subsubsection{Improving Predicate Matching}\label{subsec:tf}

\begin{figure}[t]
\centering
\includegraphics[width=0.41\textwidth]{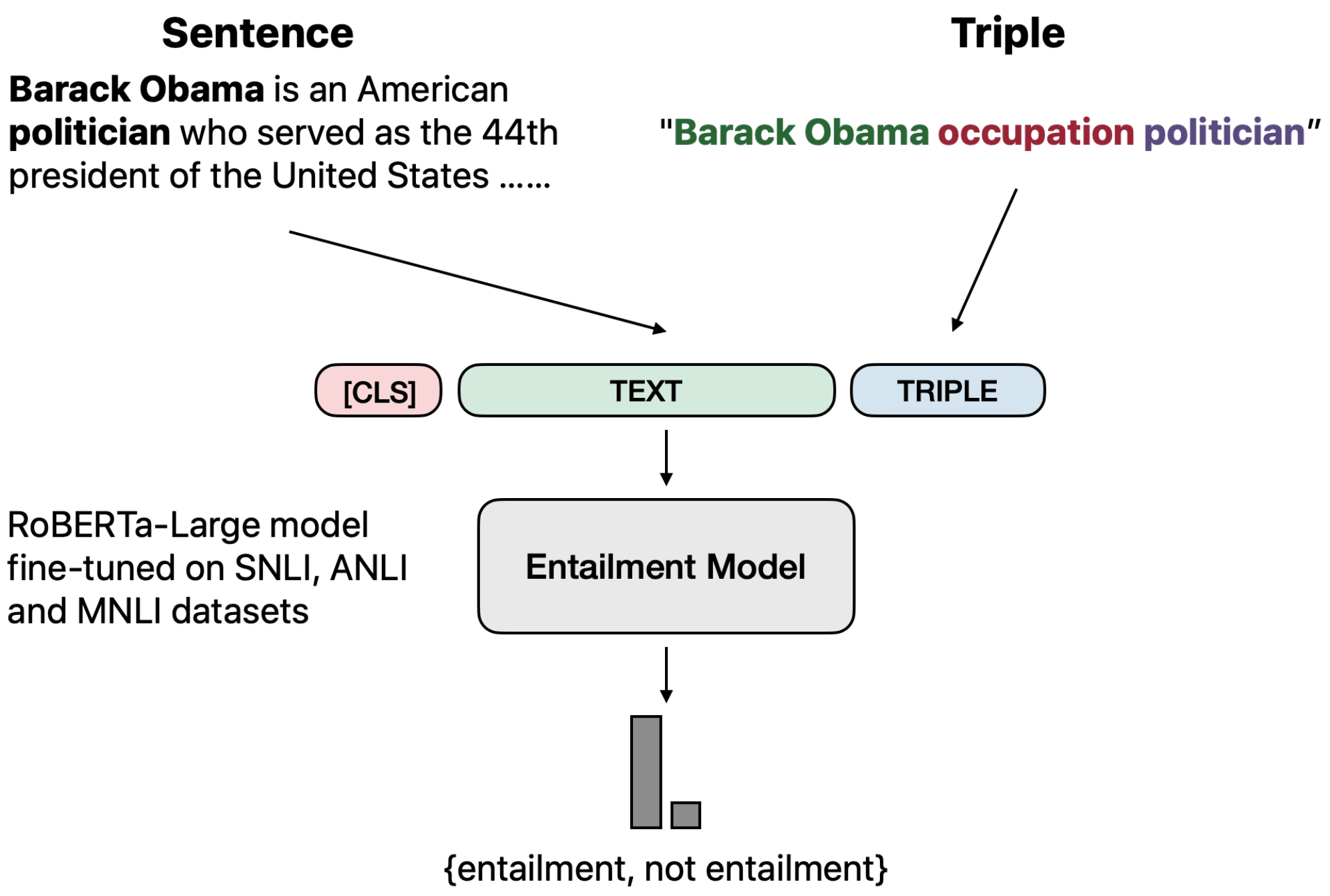}
\caption{Using the entailment model to filter the text-triple pairs.}
\label{fig:entail}
\end{figure}

The alignments generated in the previous section do not perform any verification that the text encapsulates the predicates of the triples matched to it which can lead to false matches where the triples contain information that is not present in the sentence.
To fix this, we use an entailment model to remove aligned KG triples that were not entailed by the text (See Figure~\ref{fig:entail}).
We take a RoBERTa~\cite{liu2019roberta} model\footnote{\url{https://huggingface.co/ynie/roberta-large-snli_mnli_fever_anli_R1_R2_R3-nli}} fine-tuned on natural language inference (NLI) datasets, including SNLI~\cite{bowman-etal-2015-large}, ANLI~\cite{nie2019adversarial} and MNLI~\cite{N18-1101}  and feed in a sentence and a triple as input pairs.
The entailment model produces an entailment score which predicts whether or not the sentence entails the facts described by the triple. KG triples which receive poor entailment scores are removed.

\subsubsection{Ensuring Sufficient Coverage}\label{subsec:slt}

\begin{figure}[t]
\centering
\hspace*{-0.3cm}
\includegraphics[width=0.35\textwidth]{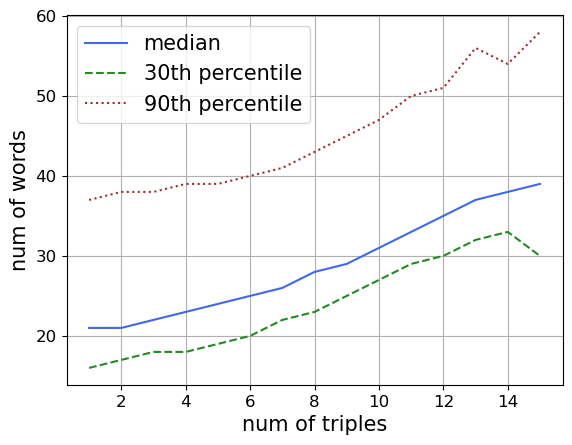}
\caption{The relationship between the number of triples and the number of words in the sentence.}
\label{fig:nw_vs_nt}
\end{figure}

The Wikipedia corpus contains a substantial number of lengthy sentences.
Many of these sentences are only covered by a limited number of matching triples because Wikidata KG covers facts in Wikipedia quite sparsely.
To mitigate this problem we remove examples where the length of the sentence appears to be longer in comparison to the number of available triples. 
To obtain a threshold we plotted the relationship between the length of sentences (measured in terms of the number of words) against the number of aligned triples (see Figure~\ref{fig:nw_vs_nt}).
As observed, there exists a roughly linear relationship between the number of aligned triples and the number of words.
We remove matches where the sentence length is greater than the $90^\text{th}$ percentile length among all examples with the same number of KG triples.
However, for single-triple examples, we set a tighter length threshold at the $30^\text{th}$ percentile length, since the vast majority of examples in our dataset contain only a single triple.

\subsubsection{Triple Augmentation}\label{subsec:ta}

A lack of coverage of the sentence can also result from KG triples either being overlooked during our construction process or not being available on Wikidata. 
We mitigate this issue by generating additional triples from a T2G model trained on the data so far. 
The generated \textit{new} triples are added as an augmentation to the original training set. 
This technique can be thought of as an analogue of back translation in neural machine translation~\cite{edunov2018understanding} where MT datasets have been augmented by using inverse models to generate new data. 
This process can be potentially repeated in iterations until the generation results converge. 
In this paper, we only run it for one iteration as a demonstration.


%% file: Sections/synthetic.tex
\subsection{Synthetic Datasets Using LLMs}\label{sec:synthetic}
As synthetic data generation using LLMs becomes widely adopted, it is interesting to understand the quality of the LLM-generated KG-text dataset using our evaluation framework.  
Due to the significant difficulty in compelling LLMs to generate KGs with canonical entity and predicate names from Wikidata, we relax our requirements and allow the triple elements to be open vocabulary. 
We prompt the LLMs to generate Wikidata style KG triples, with few-shot in-context examples as demonstration. 
We experiment with ChatGPT and Guanaco-33B~\cite{dettmers2023qlora} to generate graph from Wikipedia text with LLM instruction prompts.
The generated datasets are referred to as Guanaco-GT and ChatGPT-GT.
Given the throughput limitation of GhatGPT, we collected only 1 million examples for ChatGPT-GT.
For Guanaco-33B this was not a limitation, and we were able to generate 2.6 million examples which is on par with the size of LAGRANGE. 
More details of the LLM prompts and decoding configurations are provided in Appendix \ref{sec:llm_appendix}.


%% file: Sections/experiments.tex
\section{Experiments and Discussions}\label{sec:exps}
In this section, we first introduce our experimental setup.
Then, we show the results of three types of KG-T datasets: manually created, automatically constructed, and LLM generated.
We then present an ablation study of our proposed techniques used to create the LAGRANGE dataset.
\begin{table}[t]
\centering \small
\begin{tabular}{lcccc}
\hline
Dataset  & \#Sent. & {\hskip -0.1in}\#Tri. &{\hskip -0.1in}Avg.Tri.&{\hskip -0.1in}Avg.Words \\
\hline
WebNLG   & 35K   & 104K & 2.9 & 19.8 \\
\hline
TeKGen   & 6.3M  & 10.9M  & 1.7 & 21.3 \\
T-REx~\tablefootnote{T-REx does not split between train and dev/test. We holdout 20\% of the data for evaluation.}    & 5.0M  & 13.1M  & 2.6 & 21.8 \\
LAGRANGE & 3.0M  & 12.3M  & 4.0 & 17.9 \\
\hline
ChatGPT-GT & 1.0M  & 4.1M  & 4.2 & 17.9 \\
Guanaco-GT & 2.7M  & 15.0M  & 5.6 & 17.7 \\
\hline
\end{tabular}
\caption{Statistics of the number of sentences, number of triples, average number of triples and average number of words of the sentences. (for train splits only.)}\label{tab:datasets}
\end{table}


\begin{table*}[!htp]\centering
\small
\begin{tabular}{lcccc|ccc}\toprule
\multirow{2}{*}{Dataset} &\multicolumn{4}{c}{Cycle TGT} &\multicolumn{3}{c}{Cycle GTG} \\\cmidrule{2-8}
&BLEU-1 &BLEU-4 &ROUGE-1 &ROUGE-4 &F1 &Precision &Recall \\\midrule
WebNLG &74.68 &45.09 &79.17 &32.80 &91.42 &92.81 &91.27 \\
+10\% noise &74.19 &44.72 &78.88 &32.99 &91.25 &92.06 &90.76 \\
+20\% noise &73.41 &44.28 &78.13 &32.55 &90.55 &91.89 &89.73 \\
+30\% noise &72.69 &43.66 &77.37 &31.80 &88.36 &90.20 &87.45 \\
+40\% noise &71.21 &42.38 &76.01 &30.85 &85.50 &87.77 &84.72 \\
+50\% noise &70.71 &41.88 &75.37 &29.77 &81.18 &83.21 &81.42 \\
\bottomrule
\end{tabular}
\caption{Cyclic evaluation for WebNLG with different amount of noise.}\label{tab:webnlg-noise}
\end{table*}

\subsection{Setup}
We treat both the G2T and T2G tasks as sequence-to-sequence modeling tasks in the experiments. 
More sophisticated approaches such as~\cite{clive-etal-2022-control} can be applied under our evaluation framework, but we use a vanilla setup here for demonstration. 
To denote "subject", "predicate", "object", "qualifier", and "value" of a triple, we employ special tokens \texttt{<S>}, \texttt{<P>}, \texttt{<O>}, \texttt{<Q>}, and \texttt{<T>} respectively. 
The triples are connected by \texttt{<sep>} and serialized as a sequence. 
We fine-tune T5-large~\cite{raffel2020exploring} model on each dataset for our cycle-evaluation experiments. 
We use "\texttt{graph\_to\_text: }" as T5 prefix for G2T and "\texttt{text\_to\_graph: }" for T2G.
The models are trained with 8*A100 GPUs and the batch size is 48 for WebNLG and 192 for the others. 
We use AdamW~\cite{loshchilov2017decoupled} optimizer with the learning rate of 5e-05 and a linear decay learning rate scheduler.
The total training steps are various across different dataset: 20K for WebNLG, 50K for LLM generated datasets, and 400K for the others.
During decoding, we use the beam search size of 4.


\subsection{Metrics}
For GTG evaluation, we measure the quality of the reconstructed graph with the precision, recall, and F1 scores of triples.
For each example, we count the number of triples in the reconstructed graph that also appears in the ground-truth graph, and then calculate the scores of each example.
For TGT evaluation, we use the BLEU~\cite{papineni2002bleu} score and the ROUGE~\cite{lin2004rouge} score as metrics to evaluate the text regeneration.


\subsection{Datasets}
For evaluation, we take WebNLG v3~\cite{webnlg} as an example of human annotated KG-T dataset. 
We use LAGRANGE, TeKGen~\cite{tekgen} and T-REx~\cite{trex} are examples of KG-T datasets created by automatic alignment.
Finally, we use synthetic datasets generated by LLMs - ChatGPT-GT and Guanaco-GT by prompting.
The statistics of these datasets are shown in Table~\ref{tab:datasets}.
The test set size is 1.6K for WebNLG and 10K for the others.
In Table~\ref{tab:kgt-example}, we show an example from each of the datasets.
Additionally, we provide the statistics of various versions of our LAGRANGE dataset in Appendix~(Table~\ref{tab:datasets-full}), which will be elaborated upon in Section~\ref{section:ablation}.

\begin{table}[t]
\scriptsize
\begin{tabular}{ |p{0.6in}|p{2in}| } 
\hline
Dataset & Graph \\
\hline
LAGRANGE & 
(Kittie, has part or parts, Morgan Lander)\newline
(Morgan Lander, occupation, guitarist)\newline
(Morgan Lander, occupation, singer)\newline
(Kittie, instance of, musical group) \\ 
\hline
TeKGen &
(Kittie, has part, Morgan Lander) \\
\hline
T-REx &
(Kittie, has part, Morgan Lander)\newline
(Morgan Lander, member of, Kittie) \\
\hline
ChatGPT-GT & 
(Morgan Lander, occupation, lead vocalist)\newline
(Morgan Lander, occupation, guitarist)\newline
(Morgan Lander, band, Kittie)\newline
(Tanya Candler, occupation, bassist)\newline
(Kittie, member, Morgan Lander)\newline
(Kittie, member, Tanya Candler) \\
\hline
Guanaco-GT & 
(Morgan Lander, became, lead vocalist)\newline
(Morgan Lander, became, one of Kittie's guitarists)\newline
(Tanya Candler, completed, the band's lineup on bass)\newline
(Kittie, lineup, Morgan Lander lead vocalist one of Kittie's guitarists Tanya Candler bass) \\ 
\hline
\end{tabular}
\caption{Examples of KG triples aligned by different datasets for sentence: "\textit{Morgan Lander became the lead vocalist and one of Kittie's guitarists and Tanya Candler completed the band's lineup on bass.}". More examples are provided in Appendix \ref{sec:dataset_examples}}\label{tab:kgt-example}
\end{table}

\begin{table*}[!htp]\centering
\small
\begin{tabular}{lcccc|ccc}\toprule
\multirow{2}{*}{Dataset} &\multicolumn{4}{c}{Cycle TGT} &\multicolumn{3}{c}{Cycle GTG} \\\cmidrule{2-8}
&BLEU-1 &BLEU-4 &ROUGE-1 &ROUGE-4 &F1 &Precision &Recall \\\midrule
WebNLG &74.68 &45.09 &79.17 &32.80 &\textbf{91.42} &\textbf{91.81} &\textbf{91.27} \\\midrule
TeKGen &44.10 &28.11 &51.99 &23.09 &73.88 &74.19 &75.92 \\
T-REx &38.65 &21.39 &45.50 &16.92 &67.50 &69.80 &68.60 \\
LAGRANGE &63.38 &47.46 &67.50 &38.96 &84.33 &87.11 &84.60 \\\midrule
Guanaco-GT &\textbf{88.75} &\textbf{77.99} &\textbf{92.13} &\textbf{75.18} &41.48 &42.52 &43.56 \\
ChatGPT-GT &83.13 &68.78 &86.55 &63.03 &58.30&62.23&57.58\\
\bottomrule
\end{tabular}
\caption{Cyclic evaluation results of manually created dataset, automatically constructed datasets, and LLM generated datasets.}\label{tab:cycle-eval-main}
\end{table*}

\begin{figure*}[!htp]
\centering
\includegraphics[width=\textwidth]{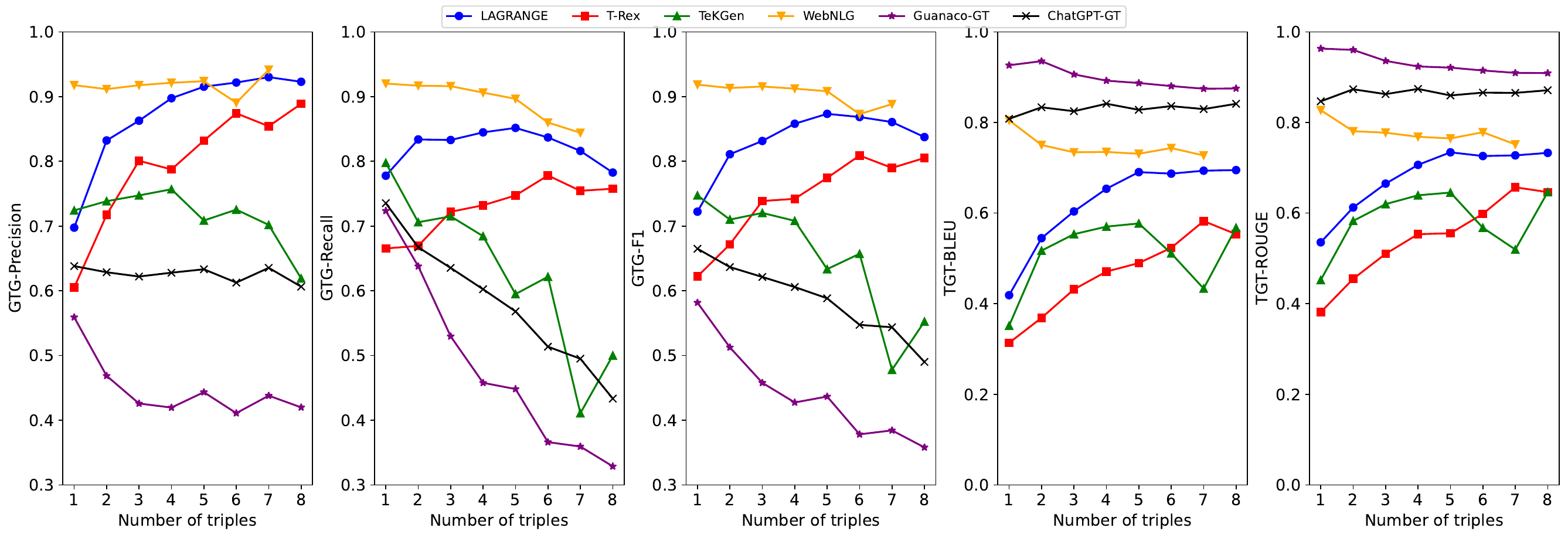}
\caption{Cyclic evaluation broke down by the number of triples.}
\label{fig:cycle-gtg-main}
\end{figure*}

\subsection{Effect of Noise in KG-T Data}
We first test our assertion that cyclic evaluation results are reflective of noise in the datasets.
To do this, we modify the WebNLG dataset by randomly inserting, deleting, or substituting triples in the examples.
We can control the level of misalignment by controlling the probability with which triples are modified\footnote{We assume that WebNLG is mostly free of noise since it was constructed manually.}.
We observe that as additional noise is introduced, the forward and reverse models trained on these noisy datasets become less accurate in both precision and recall, which confirms the "source-reference divergence" hypothesis (see Table~\ref{tab:webnlg-noise-unidirection-eval} for unidirectional evaluations and Table~\ref{tab:webnlg-noise} for cyclic evaluations). 
One might assume that neural models can deal with noise in the input, but these results indicate that the quality of models does suffer with more noise.
Further these results show that cyclic evaluation reflects the relative order of dataset quality.
Comparing unidirectional results with cyclic results we see that while the unidirectional G2T BLEU score and T2G precision demonstrated a decrease of 6.8\% and 7\%, respectively, these metrics experienced larger declines in cyclic evaluation -- with a decrease of 7.2\% for the BLEU score and and 11.2\% for precision.
This indicates the possibility that cyclic evaluations might have better resolution than unidirectional evaluations, since they can assess the effect of errors made in the unidirectional generations, on the reconstruction of the original source. 
Finally, we note again that cyclic evaluation assesses the dataset quality without the ground-truth label~(with only the text or the KG) and hence can be used to compare different datasets and ontologies.

\begin{table*}[!htp]\centering
\small
\begin{tabular}{lcccc|ccc}\toprule
\multirow{2}{*}{Dataset} &\multicolumn{4}{c}{Cycle TGT} &\multicolumn{3}{c}{Cycle GTG} \\\cmidrule{2-8}
&BLEU-1 &BLEU-4 &ROUGE-1 &ROUGE-4 &F1 &Precision &Recall \\\midrule
V0 &47.11 &31.68 &55.50 &27.74 &80.91 &82.94 &82.82 \\
V1 (V0+semantic filter) &49.69 &33.93 &57.08 &29.12 &81.16 &82.68 &83.51 \\
V2 (V1+second hop) &49.81 &34.64 &57.53 &30.00 &78.73 &80.96 &81.06 \\
V3 (V2+length filter) &62.27 &46.31 &66.27 &37.78 &82.17 &85.63 &82.30 \\
LAGRANGE (V3+augment) &\textbf{63.38} &\textbf{47.46} &\textbf{67.50} &\textbf{38.96} &\textbf{84.33} &\textbf{87.11} &\textbf{84.60} \\
\bottomrule
\end{tabular}
\caption{Ablation study.}\label{tab:cycle-ablation}
\end{table*}

\begin{table*}[!htp]\centering
\small
\begin{tabular}{lcccc|cccc}\toprule
\multirow{2}{*}{Dataset} &\multicolumn{4}{c}{Cycle TGT with TeKGen Text} &\multicolumn{4}{c}{Cycle TGT with LAGRANGE Text} \\\cmidrule{2-9}
&BLEU-1 &BLEU-4 &ROUGE-1 &ROUGE-4 &BLEU-1 &BLEU-4 &ROUGE-1 &ROUGE-4 \\\midrule
T-REx &35.74 &18.78 &42.78 &13.75 &47.88 &27.33 &50.86 &17.29 \\
TeKGen &44.10 &28.11 &51.99 &23.09 &58.25 &39.36 &62.94 &30.93 \\
V0 &40.44 &25.91 &50.02 &23.73 &61.28 &44.97 &65.55 &36.72 \\
V1 &40.73 &25.39 &48.97 &22.50 &60.74 &43.77 &64.22 &34.85 \\
V2 &41.52 &26.52 &50.01 &23.74 &62.32 &45.98 &65.93 &37.21 \\
V3 &44.22 &28.09 &52.05 &24.44 &62.27 &46.31 &66.27 &37.78 \\
LAGRANGE &\textbf{44.36} &\textbf{28.63} &\textbf{53.14} &\textbf{25.43} &\textbf{63.38} &\textbf{47.46} &\textbf{67.50} &\textbf{38.96} \\
\bottomrule
\end{tabular}
\caption{TGT results evaluated with TeKGen and LAGRANGE, respectively.}\label{tab:unified}
\end{table*}

\subsection{Main Results}\label{subsec:main_results}
The cyclic evaluation results of different datasets are shown in Table \ref{tab:cycle-eval-main}.

\textbf{Manually created dataset}. 
We can first see that WebNLG is of much higher quality: it gets the best GTG cyclic evaluation results with 91.41 F1 scores, and 74.68 BLEU-1 scores for TGT cyclic evaluation.
The results confirm that WebNLG is a well-aligned dataset.

\textbf{Automatically constructed datasets}. 
LAGRANGE gets the best results among the datasets created using automatic alignment methods, with 84.33 GTG F1 scores, 63.38 TGT BLEU-1, and 47.46 TGT BLEU-4. 
LAGRANGE demonstrates superior alignment between the text and graph compared to TeKGen and T-REx.

It is worth mentioning that LAGRANGE dataset contains a larger number of triples compared to the other datasets (Table~\ref{tab:datasets}). 
LAGRANGE achieves higher precision and recall in KG reconstruction~(GTG). 
In addition, LAGRANGE's 4-gram TGT results are better than WebNLG because WebNLG aligns multiple sentences to a set of triples. 
The introduction of sentence order in WebNLG introduces additional errors for 4-gram evaluations. 

\textbf{LLM generated datasets}. 
Guanaco-GT and ChatGPT-GT demonstrate significant superiority over the others in terms of TGT BLEU and ROUGE scores, but not in GTG evaluation.
In other words, the text is exceptionally well-reconstructed in TGT cyclic evaluation.
This can be attributed to the fact that LLMs have the ability to invent new predicates and entities, enabling them to describe relations and facts that cannot be represented by Wikidata triples.
However, this freedom also allows LLMs to generate non-existing or redundant facts and create meaningless or incoherent triples, which significantly limits the usability of the datasets.
The GTG cyclic evaluation reveals that the triples generated by LLMs are not as reproducible as those produced by LAGRANGE and other KG-grounded datasets.
We observed that Guanaco-GT performs notably better than ChatGPT-GT in TGT evaluation.
This is likely due to the Guanaco model's tendency to parse input sentences into multiple phrases, making sentence reconstruction easier.
While the ChatGPT model also suffers from this issue, its severity is relatively lower when compared to Guanaco.
This explains why ChatGPT outperforms Guanaco in GTG but not in TGT.
An example is shown in Table~\ref{tab:kgt-example}.

Finally, we visualize the cyclic evaluation results by segmenting the evaluation dataset based on the number of triples.
As illustrated in Figure \ref{fig:cycle-gtg-main}, LAGRANGE consistently surpasses TeKGen and T-REx in performance.
It is worth noting that as the number of triples increases, LAGRANGE experiences a slight decrease in recall, but its precision improves, with a more consistent F1 score, while TeKGen and LLM generated datasets decline in both the precision and recall. 
Also, it is important to highlight that while the LLM-generated datasets yield better TGT evaluation results, their GTG evaluation results are the worst.

\subsection{Ablation Study of LAGRANGE}
\label{section:ablation}

We further conducted an ablation study on our proposed techniques for constructing the LAGRANGE dataset. The results are presented in Table \ref{tab:cycle-ablation}.
As observed, all the proposed techniques consistently resulted in improvements across almost all metrics for both TGT and GTG evaluations, affirming their effectiveness.
However, there was one exception: the V2 performance for GTG. This can be attributed to an imbalance between triples based on first-hop and second-hop neighbors.
Since there are more first-hop-based triples, we observed a slight decline in precision and recall for GTG.

Meanwhile, the most significant performance gain was achieved through the \textit{length filtering} step.
This can be intuitively explained by the fact that regardless of the techniques employed, it is impossible to generate sentence segments for which corresponding triples are lacking.
Hence, the application of length filtering enhances the feasibility of sentence-graph generation.

\subsection{Unified Comparison}
Finally, we evaluate all models using the same test data for TGT evaluation. In particular, we use the TeKGen evaluation text and the LAGRANGE evaluation text, respectively. 

The results are presented in Table~\ref{tab:unified}.
It is evident that the models trained with LAGRANGE achieve the highest BLEU and ROUGE scores in both cases.
Additionally, similar to the findings in Table~\ref{tab:cycle-ablation}, the improvements resulting from our proposed techniques are also observed when evaluated with TeKGen data. 
This indicates the robustness of our cyclic evaluation approach.
Furthermore, we observe that models trained with LAGRANGE data and evaluated with TeKGen data outperform the models trained on TeKGen itself, demonstrating the effectiveness and adaptability of our evaluation methodology across different text styles.

%% file: Sections/conclusion.tex
\section{Conclusion}\label{sec:conc}

In this paper, we have addressed the alignment problem between KG and text datasets.
Our study focused on evaluating the alignment between KG triples and sentences, which lead us to propose a novel evaluation methodology that leverages the cyclic generation of the KG and of the sentences.
Using this methodology, we were able to assess the quality of alignment in existing KG-text datasets and introduced a series of techniques to enhance dataset alignment.

We also introduced the LAGRANGE dataset, showcasing significant improvements in alignment compared to existing automatically collected KG-text datasets, using cyclic evaluation. 

Finally, we created synthetic datasets using LLMs and evaluated the alignment of these LLMs-generated datasets, highlighting the advantages and disadvantages of such an approach.

To foster further research in this area, we make both the LAGRANGE dataset and the LLMs-generated dataset publicly available. We believe that these resources will serve as valuable assets for the research community to explore and advance the field of KG-text integration.

%% file: Sections/limitations.tex
\section*{Limitations}

In this paper, we propose a method to assess the alignment of KG-text datasets and a series of novel techniques to improve the alignment. There are more traditional alignment techniques that are not considered in this work (such as the ones used in \cite{trex}) since those are not the focus of this paper. LAGRANGE is based on WikiData and Wikipedia, which might not generalize well to other text domains or ontologies. Given the uneven distribution of demographics in Wikipedia corpus, LAGRANGE might inherit the bias and fairness issues (such as gender, race, occupation, etc.) from Wikipedia. Furthermore, although we have improved the alignment quality significantly, misaligned triples and phrases as well as the hallucination issue still exist to some extent. The last but not the least, triple alignment could be further improved by regenerating the sentence in each example. We do not consider that approach in this work since we prefer to keep the sentences natural.  

%% file: Sections/ethics.tex

%% file: Sections/appendix.tex
\clearpage

\onecolumn
\newpage
\appendix
\section{Appendix}
\label{sec:appendix}

\subsection{Generating an Initial Alignment between Wikipedia sentences and Wikidata triples}
\label{sec:appendix_initial_alignment}
In this section, we formally describe the methodology used to align Wikipedia entries to Wikidata KG triples, to create the initial version of LAGRANGE.

Wikipedia and Wikidata pages are associated with each other through a unique $\qidn$. For each $\qidn$ $\mathtt{q}$, we consider the set of all sentences $\sqid$ from its Wikipedia page. 
We also consider the set of all triples from the corresponding Wikidata page as $\tqid=\{(s, p, o, q, t)\}$ where $s$ denotes the subject or $\mathtt{q}$ title, $p$ denotes the predicate, $o$ denotes the object, $q$ denotes the qualifier of the predicate $p$ which is null for simple predicates, and $t$ denotes the value of the qualifier which is also null if the qualifier $q$ is null. In other words, the predicate $p$ could either be simple as in \texttt{(Albert Einstein, Occupation, Scientist)} or compound as in \texttt{(Albert Einstein, Award Received, Nobel Prize in Physics, Point in Time, 1921)}. If the predicate of a triple has a qualifier, we consider it as a compound predicate and present its qualifier and corresponding qualifier's value as well. For the sake of simplicity in our discussion, we will explain the construction process based on simple predicates. However, it's important to note that the same principles apply to compound predicates as well.

For each sentence in $\sqid$, our goal is to match as many triples as possible from $\tqid$ and neighboring Wikidata graphs to it.
Before describing the matching process, we would like to provide more details on how $\tqid$ is built.

For each $\qidn$ $\mathtt{q}$, the subject $s$ in each triple $(s, p, o) \in \tqid$ is always the title of $\mathtt{q}$ (e.g. \textit{Albert Einstein} for $\textrm{Q}_{937}$). The object $o$ on the other hand, could be 

\begin{itemize}
\item the title or an alias of another $\qidn$. As an example, \textit{Elsa Einstein} that is the title of $\textrm{Q}_{68761}$ as in \texttt{(Albert Einstein, Spouse, Elsa Einstein)}.
\item strings that are not associated to any $\qidn$.  As an example, \textit{2} as in \texttt{(Albert Einstein, Erdős number, 2)}.
\item literal values such as dates or quantities and their possible aliases. As an example, we have \texttt{(Albert Einstein, date of birth, +1879-03-14T00:00:00Z)} on Wikidata. However, on a Wikipedia sentence, this date could be expressed as \texttt{March 14, 1879} or \texttt{March 14th, 1879}. Hence, we consider different aliases of a literal in order to match more triples.
\end{itemize}

For each sentence \texttt{s} in $\sqid$, we define $\mathcal{M'}_{\mathtt{s},\mathtt{q}}$ as the set of all triples $(s, p, o) \in \tqid$ where $o$ has appeared in the sentence \texttt{s}. 
In other words, $\mathcal{M'}_{\mathtt{s},\mathtt{q}}$ denotes the set of triples matched to \texttt{s} in which objects are first-hop neighbors of $\mathtt{q}$ in Wikidata graph. 
As an example, if the sentence \texttt{s} is [\textit{Albert Einstein was born in Ulm, in the Kingdom of Württemberg in the German Empire, on 14 March 1879 into a family of secular Ashkenazi Jews.}], then $\texttt{(Albert Einstein, Place of Birth, Ulm)} \in \mathcal{M'}_{\mathtt{s},\mathrm{Q_{937}}}$ since the word \texttt{Ulm} has appeared in the sentence \texttt{s} and the entity \texttt{Ulm} (i.e., $\mathrm{Q_{3012}}$) is a first-hop neighbor of \texttt{Albert Einstein} (i.e., $\mathrm{Q_{937}}$) on Wikidata.
It is noteworthy to mention that at this stage we do not put any constraint on the predicate of matched triples.
We will later explain how our post-processing helps us to match predicates as well.

Having constructed the $\mathcal{M'}_{\mathtt{s},\mathtt{q}}$, let $\mathcal{Q}$ denote the set of all $\qidn$s in Wikidata and define $Q_{\mathcal{M'}_{\mathtt{s},\mathtt{q}}}= \{o|(s,p,o)\in {\mathcal{M'}_{\mathtt{s},\mathtt{q}}} \land \exists q'\in \mathcal{Q}: o \equiv q' \}$. In other words, for each sentence \texttt{s} in $\sqid$, $Q_{\mathcal{M'}_{\mathtt{s},\mathtt{q}}}$ denotes the set of all $\qidn$s that have appeared in $\texttt{s}$ in the original or an alias format and are first-hop neighbors of $\mathtt{q}$ in Wikidata.
In the aforementioned example where the matched triple \texttt{(Albert Einstein, Place of Birth, Ulm)} is in $\mathcal{M'}_{\mathtt{s},\mathrm{Q_{937}}}$, the $\qidn$ corresponding to \texttt{Ulm} (i.e., $\texttt{Q}_{3012}$) is also in $Q_{\mathcal{M'}_{\mathtt{s},\mathrm{Q_{937}}}}$.

\begin{figure}[h]
\centering
\includegraphics[width=0.45\textwidth]{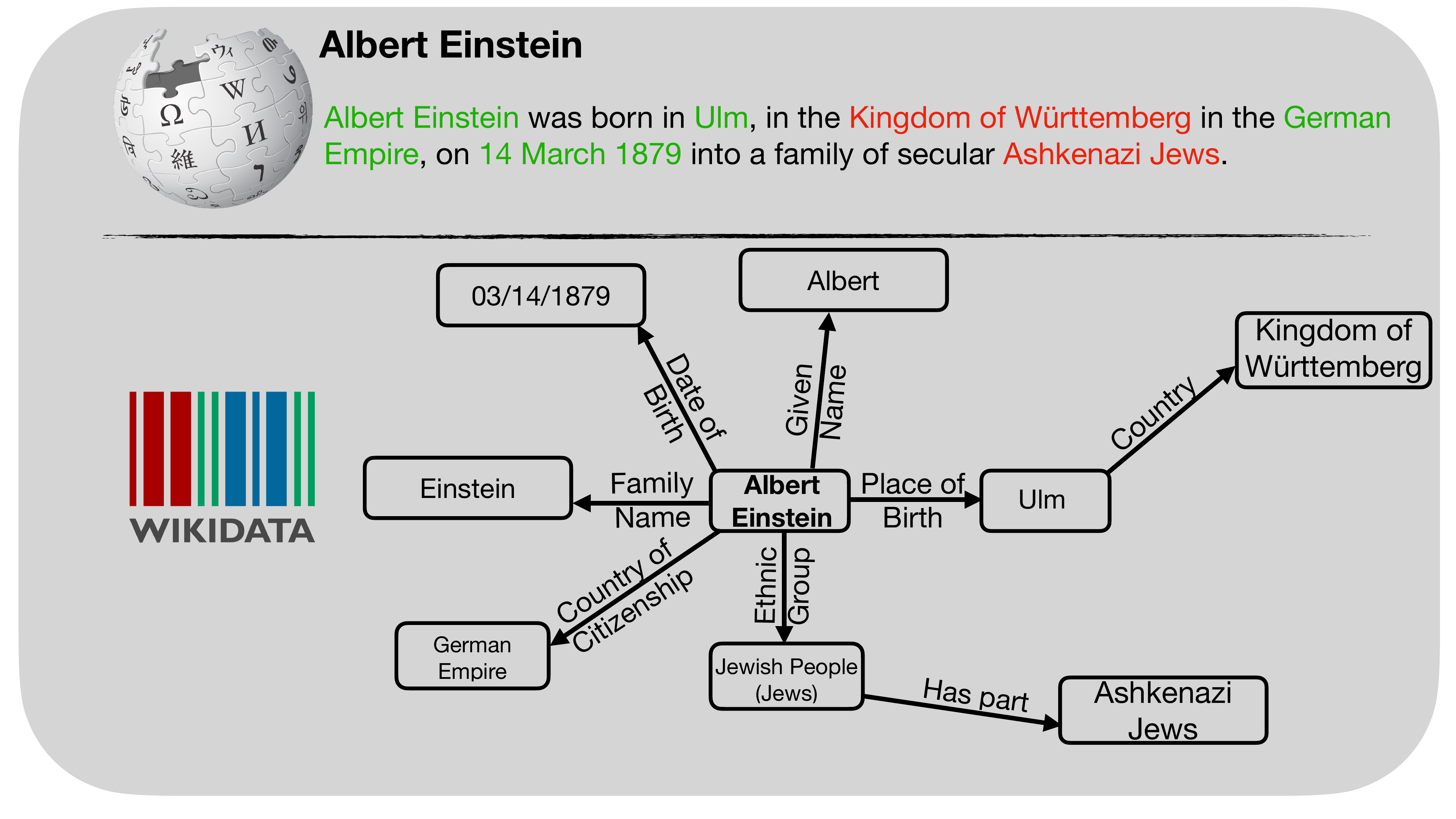}
\caption{A sample sentence and its corresponding subgraph from Albert Einstein's Wikipedia article and Wikidata. Words that are annotated in green and red are first-hop and second-hop neighbors of Albert Einstein on Wikidata, respectively.}
\label{fig:wiki_graph}
\end{figure}

In addition to $\mathcal{M'}_{\mathtt{s},\mathtt{q}}$, we define $\mathcal{M''}_{\mathtt{s},\mathtt{q}}$ as the set of all triples $(s', p', o')$ where $s'\in Q_{\mathcal{M'}_{\mathtt{s},\mathtt{q}}}$ (i.e., subject $s'$ is a $\mathrm{Q_{id}}$ and a first-hop neighbor of $\mathtt{q}$) and $o'$ has appeared in the sentence $\mathtt{s}$. As an example, consider the following sentence $\mathtt{s}$ from Albert Einstein's Wikipedia article: [\textit{Albert Einstein was born in Ulm, in the Kingdom of Württemberg in the German Empire, on 14 March 1879 into a family of secular Ashkenazi Jews.}].
Based on Albert Einstein's Wikidata graph, we have $\texttt{(Albert Einstein, Place of Birth, Ulm)} \in \mathcal{M'}_{\mathtt{s},\mathrm{Q_{937}}}$ and hence, $\mathrm{Q_{3012}}(\mathrm{Ulm}) \in Q_{\mathcal{M'}_{\mathtt{s},\mathrm{Q_{937}}}}$. Since \texttt{Ulm} is the first-hop neighbor of \texttt{Albert Einstein} and \texttt{Kingdom of Württemberg} is the first-hop neighbor of \texttt{Ulm}, then \texttt{(Ulm, Country, Kingdom of Württemberg)} is in $\mathcal{M''}_{\mathtt{s},\mathrm{Q_{937}}}$.
Figure \ref{fig:wiki_graph} shows the example sentences from Wikipedia and their corresponding subgraphs from Wikidata.
In addition, Figure \ref{fig:wiki_triples} shows the matched triples to each sentence based on the Wikidata subgraphs in Figure \ref{fig:wiki_graph}.
As seen in Figure \ref{fig:wiki_triples}, considering second-hop neighbors can significantly increase the number of matched triples and give us a richer dataset for training graph-to-text generative models.

Once we have both $\mathcal{M'}_{\mathtt{s},\mathtt{q}}$ and $\mathcal{M''}_{\mathtt{s},\mathtt{q}}$, then we build $\mathcal{M}_{\mathtt{s},\mathtt{q}}= \mathcal{M'}_{\mathtt{s},\mathtt{q}} \cup \mathcal{M''}_{\mathtt{s},\mathtt{q}}$ to denote the set of all matched triples from $\tqid$ to $\mathtt{s}$. Our raw dataset consists of 37 million sentences and 104 million triples. It is noteworthy to mention that 30 million triples in our initial dataset our based on second-hop neighbors of entities. Although our post-processing step filters out a number of these triples, we would like to emphasize on how going beyond first-hop neighbors can give us a dataset with higher coverage of information in the matched KG triples, unlike other datasets that we have mentioned them in Table \ref{tab:datasets}. 

\newpage
\subsection{Generating Synthetic Datasets from LLMs}\label{sec:llm_appendix}
We provide the LLMs prompt for synthetic datasets generation in Table \ref{tab:llm_prompts}. Since ChatGPT has a long context window, we provide few-shot examples in the prompt. The context window size of Guanaco is 2048, in order to keep the prompt concise, we show only one example in the prompt. We decode with greedy decoding so that the generation can be more stable. We do not claim these are the optimal prompts for the synthetic dataset generation task. There are rooms for tuning the prompt to improve the quality of the datasets. 

\begin{table}[!htp]
\centering \small
\begin{tabular}{|p{0.7in}|p{5in}|}
\hline
LLM  & Prompt \\
\hline
ChatGPT & Extract facts from a sentence in the form of tuples: \newline
        1. Each fact consists of a subject, a predicate, and an object. The fact might have a predicate\_attribute and an attribute\_value as well. \newline
        
        2. For predicates without any attribute, extract triples in form of (Subject, Predicate, Object) \newline
            Sentence: John is an engineer living in Chicago. \newline
            Triples: \newline
            (John, occupation, engineer) \newline
            (John, residence, Chicago) \newline
            
            Sentence: There exists an actor called Simon Pegg. \newline
            Triples: \newline
            (Simon Pegg, occupation, actor) \newline
        
        3. For predicates with an attribute, extract one tuple per attribute in the form of (Subject, Predicate, Object, Predicate\_attribute, Attribute\_value) where Predicate\_attribute is the attribute's name, and Attribute\_value is the attribute's value. \newline
            Sentence: John started working at Apple since 2008. \newline
            Triples: \newline
            (John, employer, Apple, start time, 2008) \newline
            
            Sentence: Sara and Bob got divorced at 2012. \newline
            Triples: \newline
            (Sara, spouse, Bob, end time, 2012) \newline
            (Sara, spouse, Bob, end cause, divorce) \newline

    Sentence: \{sentence\} \newline
    Triples: \newline
 \\
\hline
Guanaco-33B &  Below is an instruction that describes a task, paired with an input that provides further context. Write a response that appropriately completes the request. \newline
\#\#\# Instruction: \newline
Extract WikiData knowledge graph triples from the following sentence. For example,\newline
Sentence: Obama was elected over Republican nominee John McCain in the presidential election and was inaugurated on January 20, 2009.\newline
Triples: \newline
(Obama, elected over, John McCain) \newline
(Obama, inaugurated on, January 20, 2009) \newline
(Obama, election type, presidential election) \newline
(Obama, nominee, Republican nominee) \newline
(Obama, election date, January 20, 2009) \newline
\#\#\# Human:  Sentence: \{sentence\} \newline
\#\#\# Assistant: \newline\\
\hline
\end{tabular}
\caption{The LLM prompts used for synthetic datasets generation.}\label{tab:llm_prompts}
\end{table}



\newpage
\subsection{Comparison between construction techniques of different datasets}
Here we also contrast other approaches to create KG-T datasets with the method we used to create LAGRANGE in Table~\ref{tab:data-features}. 
WebNLG \citep{webnlg}) is a small scale manually created dataset and is thus of high quality but has a limited ontology.  TEKGEN\citep{tekgen} is a recently developed dataset that aligns Wikidata triples to the sentences in the first section of the corresponding Wikipedia articles, based on the presence of the triple object or its aliases in the sentence.  However, does not assess whether predicates in the graph match appropriately with the sentence.
KGPT\citep{kgpt}, on the other hand, relies on unigram overlaps between sentences and graphs using Wikipedia hyperlinks, which can lead to missing matches for non-hyperlinked entities.
GenWiki \citep{jin2020genwiki} is similar to KGPT, but smaller, and aligns triples and sentences based on entity set overlaps without predicate matching.
In contrast, T-REx utilizes predicate linker and coreference resolution to match triples to sentences, but can miss matches when the predicate is semantically entailed in a sentence but not explicitly mentioned. WikiGraphs \citep{wikigraphs}, unlike T-REx and other previously mentioned datasets, matches entire Wikipedia articles instead of individual sentences to Freebase KG \citep{freebase}, and it does not rely on predicate matching.
\begin{table}[!htp]
\caption{Comparing different attributes of LAGRANGE with other datasets.}
\label{sample-table}
\vskip 0.15in
\begin{center}
\scriptsize
\begin{sc}
\begin{tabular}{lccccccr}
\toprule
Property & LAGRANGE & TEKGEN & KGPT & GenWiki & WikiGraphs & T-REx & WebNLG  \\
\midrule

human made    & $\times$& $\times$ & $\times$ & $\times$ & $\times$ & $\times$ & $\surd$ \\
second-hop coverage    & $\surd$& $\textbf{}\times$ & $\times$ & $\times$ & $\times$ & $\times$ & $\times$ \\
Non-hyperlinked annotation    & $\surd$& $\surd$ & $\times$ & $\surd$ & $\surd$ & $\surd$ & $\times$ \\
predicate matching    & $\surd$& $\times$ & $\times$ & $\times$ & $\times$ & $\surd$ & $\surd$ \\

semantic alignment    & $\surd$& $\times$ & $\times$ & $\times$ & $\times$ & $\times$ & $\surd$ \\
\bottomrule
\end{tabular}
\end{sc}
\end{center}
\vskip -0.1in
\label{tab:data-features}
\end{table}

\subsection{Datasets Statistics}\label{sec:dataset_details}

\begin{table}[!htp]
\centering \small
\begin{tabular}{lcccc}
\hline
Dataset  & \#Sent. & \#Tri. & \#Tri./Sent. & \#Words \\
\hline
WebNLG   & 35K   & 104K     & 2.96 & 19.83 \\
\hline
TeKGen   & 6.3M  & 10.9M  & 1.73 & 21.26 \\
T-Rex    & 5.0M  & 13.1M  & 2.61 & 21.83 \\
V0       & 5.2M  & 14.2M  & 2.70 & 20.38 \\
V1(V0+semantic filter)       & 4.2M & 10.0M   & 2.40 & 20.78 \\
V2(V1+second hop triples)      & 4.3M & 13.0M  & 3.00 & 20.77 \\
V3(V2+length filtering)       & 3.0M  & 11.0M  & 3.59 & 17.90 \\
LAGRANGE(V3+augment)          & 3.0M  & 12.3M  & 4.02 & 17.90 \\
\hline
ChatGPT-GT       & 1.0M  & 4.1M  & 4.17 & 17.90 \\
Guanaco-GT      & 2.7M  & 15.0M  & 5.59 & 17.72 \\
\hline
\end{tabular}
\caption{Statistics of the number of sentences, number of triples, and number of triples per sentence.}\label{tab:datasets-full}
\end{table}

\newpage
\subsection{Dataset Examples}\label{sec:dataset_examples}
\begin{table}[!htp]
\scriptsize
\begin{tabular}{ |p{1.2in}|p{0.9in}|p{3.3in}| } 
\hline
Text & Dataset & Graph \\
\hline
\multirow{5}{1.2in}{Uncommon Women and Others (1977), is the first play by noted 20th-century American playwright Wendy Wasserstein.} 
& LAGRANGE & 
(Uncommon Women and Others, instance of, play)\newline
(Uncommon Women and Others, author, Wendy Wasserstein)\newline
(Wendy Wasserstein, occupation, playwright) \\ 
\cline{2-3}
& TeKGen & 
(Uncommon Women and Others, author, Wendy Wasserstein) \\ 
\cline{2-3}
& T-REx & 
(Wendy Wasserstein, occupation, playwright)\newline
(Wendy Wasserstein, country of citizenship, United States of America)\newline
(Uncommon Women and Others, author, Wendy Wasserstein) \\ 
\cline{2-3}
& ChatGPT-GT & 
(Wendy Wasserstein, notable for, 20th-century American playwright)\newline
(Uncommon Women and Others, type, play)\newline
(Uncommon Women and Others, year, 1977)\newline
(Uncommon Women and Others, author, Wendy Wasserstein)\newline
(Uncommon Women and Others, first, true) \\
\cline{2-3}
& Guanaco-GT & 
(Uncommon Women and Others, written by, Wendy Wasserstein)\newline
(Uncommon Women and Others, first play, 1977)\newline
(Uncommon Women and Others, 20th-century, American playwright Wendy Wasserstein)\newline
(Uncommon Women and Others, noted, 20th-century American playwright Wendy Wasserstein)\newline
(Uncommon Women and Others, play, Uncommon Women and Others) \\ 
\hline\multirow{3}{1.2in}{Stewart's Restaurants are classic 1950s style fast-food restaurants located throughout the United States.} 
& LAGRANGE & 
(Stewart's Restaurants, instance of, restaurant)\newline
(Stewart's Restaurants, country, United States of America) \\ 
\cline{2-3}
& TeKGen &
(Stewart 's Restaurants, country, United States) \\
\cline{2-3}
& T-REx &
(Stewart's Restaurants, country, United States of America) \\
\cline{2-3}
& ChatGPT-GT & 
(Stewart's Restaurants, type, fast-food restaurant)\newline
(Stewart's Restaurants, style, 1950s)\newline
(Stewart's Restaurants, location, United States) \\
\cline{2-3}
& Guanaco-GT & 
(Stewart's Restaurants, style, 1950s)\newline
(Stewart's Restaurants, location, United States)\newline
(Stewart's Restaurants, type, fast-food)\newline
(Stewart's Restaurants, date, classic)\newline
(Stewart's Restaurants, location, throughout) \\ 
\hline\multirow{3}{1.2in}{The 2009 CAF Champions League is the 45th edition of Africa's premier club football tournament organized by the Confederation of African Football (CAF), and the 13th edition under the current CAF Champions League format.}
& LAGRANGE &
(Confederation of African Football, operating area, Africa)\newline
(2009 CAF Champions League, organizer, Confederation of African Football)\newline
(2009 CAF Champions League, sport, association football)\newline
(Confederation of African Football, short name, CAF)\newline
(2009 CAF Champions League, point in time, 2009)\newline
(2009 CAF Champions League, sports season of league or competition, CAF Champions League)
\\
\cline{2-3}
& TeKGen &
(2009 CAF Champions League, point in time, 00  2009)
\\
\cline{2-3}
& T-REx &
(2009 CAF Champions League, instance of, CAF Champions League)\newline
(CAF Champions League, sport, association football)\newline
(2009 CAF Champions League, sport, association football)\newline
(Confederation of African Football, sport, association football)
\\
\cline{2-3}
& ChatGPT-GT &
(2009 CAF Champions League, edition number, 45th)\newline
(2009 CAF Champions League, tournament name, Africa's premier club football tournament)\newline
(2009 CAF Champions League, organizer, Confederation of African Football)\newline
(2009 CAF Champions League, tournament format, CAF Champions League)\newline
(2009 CAF Champions League, current edition number, 13th)
\\
\cline{2-3}
& Guanaco-GT &
(2009 CAF Champions League, edition, 45th)\newline
(2009 CAF Champions League, format, CAF Champions League)\newline
(2009 CAF Champions League, format, 13th edition)\newline
(2009 CAF Champions League, organization, Confederation of African Football)\newline
(2009 CAF Champions League, tournament, Africa's premier club football tournament)\newline
(2009 CAF Champions League, tournament, football tournament)
\\
\hline
\end{tabular}
\caption{Additional examples of KG-Text datasets.}\label{tab:additional}
\end{table}

\newpage
\subsection{Cyclic Generation Examples}
\begin{table}[!htp]
\scriptsize
\begin{tabular}{ |p{1.2in}|p{0.8in}|p{1.8in}|p{1.7in}| } 
\hline
Input Text & Dataset & Predicted Graph & Reconstructed Text \\
\hline
\multirow{5}{1.2in}{Toba Domain was centered on Toba Castle in what is now the city of Toba.} 
& LAGRANGE & 
(Toba Domain,  capital,  Toba)\newline
(Toba,  instance of,  city)\newline
(Toba Domain,  headquarters location,  Toba Castle)&
Toba Domain was centered on Toba Castle in what is now the city of Toba.
\\\cline{2-4}
& TeKGen & 
(Toba Domain, headquarters location, Toba Castle) &
Toba Domain was centered on Toba Castle in what is now the city of Toba in Nagano Prefecture.
\\\cline{2-4}
 & T-REx & 
(Toba Castle, located in the administrative territorial entity, Toba) &
Toba Castle (Serbian: \textbackslash u0430 \textbackslash u043c \textbackslash u043e \textbackslash u043a \textbackslash u043e \textbackslash u0430, Zamek Toba) is a castle in Toba, Serbo-Croatian Voivodeship, Poland.
\\\cline{2-4}
 & ChatGPT-GT & 
(Toba Domain, center, Toba Castle)\newline(Toba Domain, location, city of Toba) &
Toba Domain was centered on Toba Castle in the city of Toba.
\\ \cline{2-4}
& Guanaco-GT &
(Toba Domain, centered on, Toba Castle)\newline(Toba Domain, in, what is now the city of Toba)\newline(Toba Domain, was, Toba Castle)\newline(Toba Domain, was, what is now the city of Toba)\newline(Toba Domain, was, centered on Toba Castle) &
Toba Domain was centered on Toba Castle in what is now the city of Toba.
 \\\hline
 \multirow{5}{1.2in}{Mark Sandrich (born Mark Rex Goldstein; October 26, 1900 - 2013 March 4, 1945) was an American film director, writer, and producer.} 
& LAGRANGE & 
(Mark Sandrich, date of death, March 4, 1945)\newline
(Mark Sandrich, occupation, film director)\newline
(Mark Sandrich, date of birth, October 26, 1900) \newline
(Mark Sandrich, occupation, film producer) \newline
(Mark Sandrich, date of death, 1945)&
Mark Sandrich (October 26, 1900 - 2013 March 4, 1945), born Mark Sandritz, was an American film director and producer.
\\\cline{2-4}
& TeKGen & 
(Mark Sandrich, date of death, 04 March 1945)\newline
(Mark Sandrich, occupation, Film director)\newline
(Mark Sandrich, date of birth, 26 October 1900) &
Mark Sandrich (26 October 1900 - 4 March 1945) was a German film director. 
\\\cline{2-4}
 & T-REx & 
(Mark Sandrich, country of citizenship, United States of America)\newline
(Mark Sandrich, occupation, film director) &
Mark Sandrich is an American film director. 
\\\cline{2-4}
 & ChatGPT-GT & 
(Mark Sandrich, occupation, film director)\newline
(Mark Sandrich, occupation, writer)\newline
(Mark Sandrich, occupation, producer)\newline
(Mark Sandrich, birth name, Mark Rex Goldstein)\newline
(Mark Sandrich, nationality, American) &
Mark Sandrich (born Mark Rex Goldstein; July 7, 1939) is an American film director, writer, and producer.
\\ \cline{2-4}
& Guanaco-GT &
(Mark Sandrich, born, Mark Rex Goldstein)\newline
(Mark Sandrich, died, March 4 1945)\newline
(Mark Sandrich, profession, film director)\newline
(Mark Sandrich, profession, writer)\newline
(Mark Sandrich, profession, producer)\newline
(Mark Sandrich, birth date, October 26 1900) &
Mark Sandrich (born Mark Rex Goldstein; October 26, 1900 - 2013 March 4, 1945) was an American film director, writer, and producer.
 \\\hline
\end{tabular}
\caption{Generations of TGT cyclic evaluation.}\label{tab:tgt-generation-example}
\end{table}